\documentclass{article}

\usepackage{PRIMEarxiv}

\usepackage[utf8]{inputenc} % allow utf-8 input
\usepackage[T1]{fontenc}    % use 8-bit T1 fonts
\usepackage{hyperref}       % hyperlinks
\usepackage{url}            % simple URL typesetting
\usepackage{booktabs}       % professional-quality tables
\usepackage{amsfonts}       % blackboard math symbols
\usepackage{nicefrac}       % compact symbols for 1/2, etc.
\usepackage{microtype}      % microtypography
\usepackage{amsmath}        % for \lVert, \rVert, etc.
\usepackage{fancyhdr}       % header
\usepackage{graphicx}       % graphics
\graphicspath{{media/}}     % organize your images and other figures under media/ folder
\usepackage[ruled,vlined]{algorithm2e}
\usepackage{tabularx}

\begin{document}

%% Title
\title{Hybrid LSTM and PPO Networks for Dynamic Portfolio Optimization

}

\author{
  Jun Kevin \\
  Universitas Pelita Harapan \\
  Jakarta\\
  Indonesia\\
  \texttt{01679240002@student.uph.edu} \\
   \And
  Pujianto Yugopuspito \\
  Universitas Pelita Harapan \\
  Jakarta\\
  Indonesia\\
  \texttt{yugopuspito@uph.edu} \\
}

\maketitle

\begin{abstract}
This paper introduces a hybrid framework for portfolio optimization that fuses Long Short-Term Memory (LSTM) forecasting with a Proximal Policy Optimization (PPO) reinforcement learning strategy. The proposed system leverages the predictive power of deep recurrent networks to capture temporal dependencies, while the PPO agent adaptively refines portfolio allocations in continuous action spaces, allowing the system to anticipate trends while adjusting dynamically to market shifts. Using multi-asset datasets covering U.S. and Indonesian equities, U.S. Treasuries, and major cryptocurrencies from January 2018 to December 2024, the model is evaluated against several baselines, including equal-weight, index-style, and single-model variants (LSTM-only and PPO-only). The framework's performance is benchmarked against equal-weighted, index-based, and single-model approaches (LSTM-only and PPO-only) using annualized return, volatility, Sharpe ratio, and maximum drawdown metrics, each adjusted for transaction costs. The results indicate that the hybrid architecture delivers higher returns and stronger resilience under non-stationary market regimes, suggesting its promise as a robust, AI-driven framework for dynamic portfolio optimization.
\end{abstract}

% keywords can be removed

\section{Introduction}

Portfolio optimization lies at the core of contemporary investment management, focusing on how capital can be distributed across multiple asset classes (such as equities, bonds, and digital assets) to achieve an optimal balance between return and risk. However, as global markets grow more intricate and interconnected, and as new instruments like cryptocurrencies emerge, classical frameworks such as Markowitz’s Modern Portfolio Theory (MPT) have shown notable shortcomings \cite{harris2024bitcoin,cui2023multi}. Although MPT provides a crucial theoretical foundation, its reliance on assumptions of normal return distributions and fixed correlations often proves unrealistic amid the turbulence and regime shifts characteristic of modern financial environments \cite{jabbar2024comprehensive,bedoui2023hybrid}.

The inability of static, linear frameworks to capture non-linear dynamics has accelerated the shift toward data-driven, adaptive approaches. Deep learning, with its capacity to uncover complex dependencies in large-scale time series, offers a powerful alternative \cite{bao2017deep,paiva2019decision,bouteska2024cryptocurrency}. Among various deep learning models, Long Short-Term Memory (LSTM) networks have demonstrated notable effectiveness in capturing and predicting temporal dynamics within financial time series \cite{zuniga2023downtrend,li2020hybrid}. Nevertheless, despite their predictive strength, LSTMs are not inherently designed to determine optimal portfolio allocations or adjust investment actions in a sequential and adaptive manner.

Deep Reinforcement Learning (DRL) addresses this limitation by enabling agents to learn dynamic allocation strategies through continuous interaction with market environments \cite{ye2020sarl,choi2024expert}. Among DRL algorithms, Proximal Policy Optimization (PPO) stands out for its stability and efficiency in continuous action spaces, making it ideal for portfolio control \cite{wang2021deeptrader,yue2022mdpautoencoder}. Yet, DRL agents can be data-hungry and unstable under volatile conditions if deprived of predictive priors \cite{james2023entropy}.

This paper proposes a hybrid portfolio optimization framework that integrates LSTM-based return forecasting with PPO-driven allocation. By combining predictive foresight with adaptive decision-making, the model aims to achieve superior risk-adjusted returns under realistic, multi-asset conditions. We evaluate its performance against conventional baselines (Index Fund and Equal-Weight) and single-model counterparts (LSTM-only, PPO-only), evaluated through core performance indicators including annualized return, risk volatility, Sharpe ratio, and maximum drawdown. The results highlight that hybrid AI architectures can serve as a robust and flexible foundation for modern portfolio management \cite{lv2023dynamic,bedoui2023hybrid}.

\section{Related Work}

\subsection{Portfolio Optimization} Portfolio optimization has evolved significantly since Markowitz introduced the Modern Portfolio Theory (MPT), which formalized the risk–return trade-off through mean–variance analysis. Although foundational, MPT assumes static correlations and normally distributed returns—assumptions often invalid in dynamic markets characterized by rare events and non-linear dependencies \cite{jaquart2022mlcrypto,millea2023drlandhrp}. Recent studies extend this framework by accounting for uncertainty, transaction costs, and high-dimensionality. For instance, Lv et al. \cite{lv2023dynamic} developed a dynamic portfolio model incorporating diffusion and uncertainty arising from diffusion and abrupt price jumps in stock and cryptocurrency markets, finding that investors behave asymmetrically under upward and downward price shocks. Similarly, James and Menzies \cite{james2023entropy} investigated diversification and collective dynamics in crypto portfolios, showing that correlation structures fluctuate drastically during crises such as exchange collapses, undermining the benefits of diversification.

These advances highlight the shift from static mean–variance optimization toward adaptive and robust frameworks capable of handling volatility clustering and structural breaks. Machine learning–based models now dominate recent approaches, providing non-parametric flexibility in modeling uncertainty and regime-switching behaviors \cite{yue2022mdpautoencoder,wang2024tnnls}.

\subsection{LSTM (Long Short-Term Memory) Forecasting in Finance} As financial markets generate increasingly high-frequency and non-stationary data, deep learning approaches (especially Long Short-Term Memory (LSTM) networks) have exhibited remarkable effectiveness in modeling and forecasting complex financial time series \cite{harris2024bitcoin,zuniga2023downtrend}. Unlike ARIMA or GARCH models, LSTMs capture long-term dependencies without strong distributional assumptions \cite{li2024contrastive}. AlMadany et al. \cite{almadany2024forecasting} compared LSTM against classical statistical models and hybrid EGARCH-LSTM variants, finding deep models produced lower forecasting errors across ten major cryptocurrencies. Seabe et al. \cite{seabe2023forecasting} confirmed this by showing that Bi-LSTM outperformed standard LSTM and GRU architectures in predicting Bitcoin, Ethereum, and Litecoin prices, achieving mean absolute percentage errors below 5\%.

Hybrid and enhanced LSTM frameworks further improve forecasting performance. For instance, integrating volatility models (GARCH-LSTM, EGARCH-LSTM) has been shown to stabilize forecasts in high-volatility regimes \cite{lahmiri2021deep}. Nevertheless, as emphasized by AlMadany et al. \cite{almadany2024forecasting} and Seabe et al. \cite{seabe2023forecasting}, deep networks remain sensitive to overfitting and data noise, challenges that motivate coupling them with reinforcement learning for real-time adaptability.

Formally, given a return sequence $r_t$, the LSTM predicts $\hat r_{t+1}=f_{\theta}\big(r_{t-L:t}\big)$, 
where $f_{\theta}$ represents the recurrent mapping with parameters $\theta$ trained to minimize loss 
$\mathcal{L}=\sum_t (\hat r_t-r_t)^2$.

\begin{figure}[h!]
  \centering
  \includegraphics[width=0.95\linewidth]{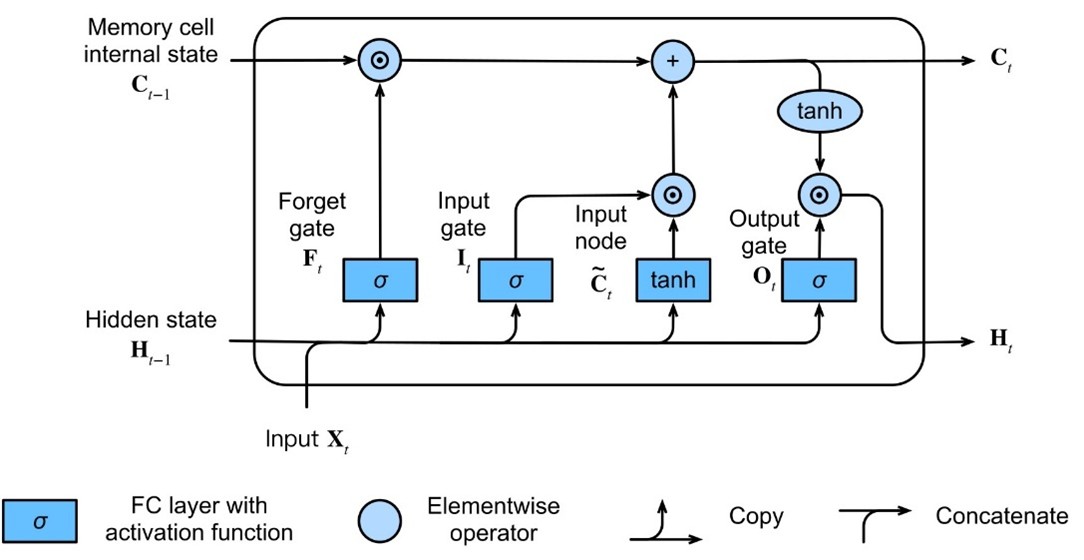}
  \caption{LSTM cell architecture illustrating the input, forget, and output gates, 
  which regulate information flow between the cell state ($C_t$) and hidden state ($H_t$).}
  \label{fig:lstm-architecture}
\end{figure}

The predicted returns can later serve as signals for reinforcement-based allocation.

\subsection{Deep Reinforcement Learning in Portfolio Management} Reinforcement learning (RL) formulates the portfolio allocation problem as a Markov Decision Process (MDP), where an agent iteratively learns the optimal portfolio weights $w_t$ that maximize cumulative rewards—typically represented as the expected logarithmic return adjusted for risk. Vetrin and Koberg \cite{vetrin2024reinforcement} formalized this paradigm by applying deep RL to trading strategy optimization, showing that algorithms such as Deep Q-Networks (DQN), Deep Deterministic Policy Gradient (DDPG), and Proximal Policy Optimization (PPO) can dynamically manage asset allocations. Similarly, Zuniga et al. \cite{zuniga2023downtrend} demonstrated that RL-based systems incorporating transaction costs and borrowing constraints deliver enhanced stability and higher cumulative performance compared to conventional optimization methods.

Recent studies extend this to the cryptocurrency domain, where volatility, liquidity constraints, and dynamic asset compositions challenge static portfolio strategies. Wang et al. \cite{wang2021deeptrader} introduced a deep reinforcement learning framework capable of managing portfolios with a changing number of assets, allowing adaptive reallocation as instruments enter or exit the market. Sadighian \cite{sadighian2020marketmaking} further advanced reinforcement learning for cryptocurrency trading by employing an event-driven environment and adaptive reward functions, improving trading stability and profitability under highly volatile market conditions.

Mathematically, the RL objective is to find a policy $\pi(a_t\mid s_t)$ that maximizes expected discounted rewards:
$$J(\pi)=\mathbb{E}_{\pi} \Big[ \sum_{t=0}^{T} \gamma^t R_t \Big],$$
where $R_t=\log (1+r_t^{\top} w_t) - \lambda \lVert \Delta w_t \rVert_1$ accounts for both return and transaction cost \cite{almadany2024forecasting}.

PPO improves stability through \textbf{clipped policy updates}:
\begin{equation}
\mathcal{L}^{\operatorname{CLIP}}(\theta)=\mathbb{E}_t\big[\min\big(r_t(\theta) \hat A_t, \operatorname{clip}(r_t(\theta),1-\epsilon,1+\epsilon)\hat A_t\big)\big], \label{eq:ppo_clip}
\end{equation}
where $r_t(\theta)=\tfrac{\pi_{\theta}(a_t\mid s_t)}{\pi_{\theta_{\text{old}}}(a_t\mid s_t)}$ is the probability ratio between the new and old policies, and $\mathbf{\hat A_t}$ is the \textbf{Advantage Estimate}. The function $\mathbf{\operatorname{clip}(\cdot)}$ restricts this ratio within a small interval $[1-\epsilon, 1+\epsilon]$ to prevent destructive large policy updates \cite{soleymani2021deeppocket}.

\subsection{Alternative Algorithmic Combinations for Portfolio Optimization} Beyond traditional neural or reinforcement learning models, recent studies explore multi-algorithm combinations to enhance portfolio robustness and adaptability. Jaquart et al. \cite{jaquart2022mlcrypto} evaluated a suite of machine learning approaches (including random forest and gradient boosting) for cryptocurrency trading, showing that ensemble long-short strategies achieved Sharpe ratios above 3.0 after transaction costs, outperforming market benchmarks. Similarly, Lahmiri and Bekiros \cite{lahmiri2021deep} demonstrated that deep feed-forward neural networks optimized with the Levenberg–Marquardt algorithm improved high-frequency Bitcoin forecasts, highlighting the role of optimization techniques in predictive accuracy.

Other hybrid frameworks combine model interpretability and dynamic risk control. Millea and Edalat \cite{millea2023drlandhrp} merged deep reinforcement learning with hierarchical risk parity (HRP/HERC) allocation, where a high-level DRL agent adaptively selected among low-level risk-based models, achieving superior robustness across asset classes. Yue et al. \cite{yue2022mdpautoencoder} proposed an MDP-based deep reinforcement framework (SwanTrader) integrating autoencoder-based feature augmentation and a risk-aware actor–critic agent optimized under the omega ratio, which proved resilient during the COVID-19 market turbulence.

These approaches collectively indicate a shift toward composite learning architectures (where predictive, structural, and decision-making algorithms operate jointly) to improve stability and generalization under non-stationary financial regimes.

\section{Methodology}

\subsection{Research Methodology}

\begin{figure}[h!]
  \centering
  \includegraphics[width=0.95\linewidth]{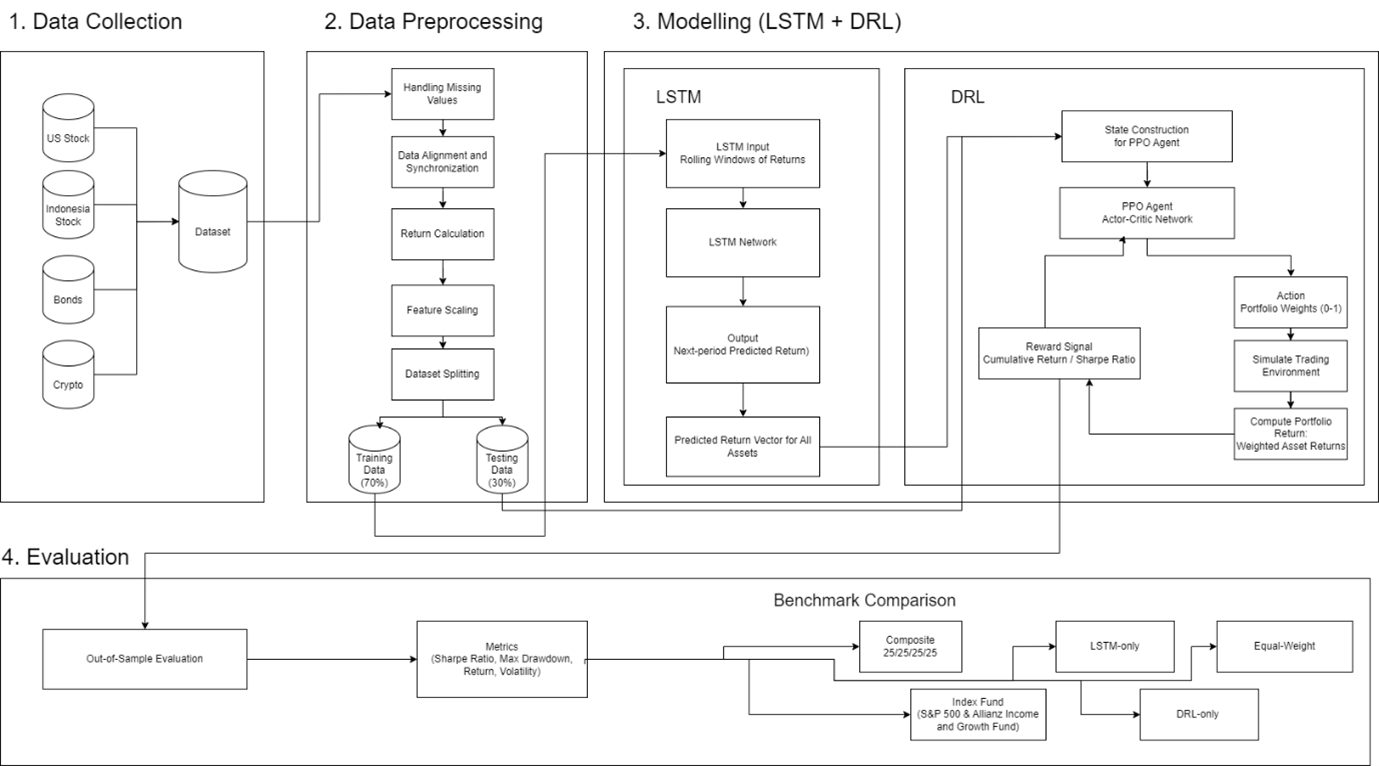}
  \caption{Research pipeline integrating data preparation, hybrid modeling (LSTM + PPO), and evaluation.}
  \label{fig:research-flow}
\end{figure}

This study employs a multi-stage, data-driven framework that integrates deep forecasting and sequential decision-making for robust portfolio optimization (Figure~\ref{fig:research-flow}). The process comprises four stages: data collection, data preprocessing, hybrid modeling (LSTM + PPO), and evaluation.

First, multi-asset data are collected from four major classes—U.S. equities, Indonesian equities, government bonds, and cryptocurrencies—to capture diverse market dynamics. Next, preprocessing ensures data consistency through missing-value imputation, time alignment, log-return computation, and normalization. The dataset is then chronologically divided into 70\% training and 30\% testing subsets to maintain strict out-of-sample evaluation.

In the modeling stage, an LSTM forecaster learns temporal dependencies to predict next-period returns, while a PPO agent translates these predictive signals into dynamic, transaction-aware portfolio weights. The agent optimizes a clipped-surrogate reward based on cumulative return and Sharpe ratio, iteratively refining allocations through interaction with a simulated trading environment. Finally, model performance is compared against baselines—including Equal-Weight, Index, and single-model variants—using annualized return, volatility, Sharpe ratio, and maximum drawdown.

\subsection{Data Collection and Preprocessing}

The dataset integrates four asset classes—U.S. equities (Nasdaq-100), Indonesian equities (IDX30), U.S. 10-Year Treasury yields (\^TNX), and the top ten cryptocurrencies by market capitalization (Table~\ref{tab:top10-crypto}). This multi-market composition enables the model to generalize across heterogeneous volatility structures and correlation regimes.

\begin{table}[h!]
\centering
\caption{Top 10 cryptocurrencies by market cap as of May 1st, 2025 \cite{coinmarketcap2025}.}
\label{tab:top10-crypto}
\begin{tabular}{lllr}
\toprule
Rank & Cryptocurrency & Ticker & Market Cap (USD) \\
\midrule
1  & Bitcoin & BTC & 1.9T \\
2  & Ethereum & ETH & 216B \\
3  & XRP & XRP & 129B \\
4  & Binance Coin & BNB & 85B \\
5  & Solana & SOL & 77B \\
6  & Dogecoin & DOGE & 26B \\
7  & Cardano & ADA & 24B \\
8  & TRON & TRX & 23B \\
9  & Toncoin & TON & 23B \\
10 & Polkadot & DOT & 22B \\
\bottomrule
\end{tabular}
\end{table}

All data were sourced from Yahoo Finance (2018–2024) at daily frequency and resampled to weekly intervals. Missing values were forward-filled, trading calendars synchronized, and closing prices transformed into log-returns:
\begin{equation}
r_t = \ln\!\left(\frac{P_t}{P_{t-1}}\right),
\end{equation}
which improves stationarity and preserves proportional price changes. Each return series was standardized using z-score normalization:
\begin{equation}
z_t = \frac{r_t - \mu}{\sigma},
\end{equation}
to ensure consistent scaling across assets. All experiments were implemented in Python using \texttt{pandas}, \texttt{NumPy}, and \texttt{scikit-learn} for data handling, \texttt{TensorFlow} for LSTM modeling, and \texttt{Stable-Baselines3} for PPO training. This unified computational pipeline ensures full reproducibility.

\subsection{LSTM Forecasting Module}

The Long Short-Term Memory (LSTM) module predicts one-step-ahead weekly returns for each asset using past return sequences. Each univariate LSTM captures temporal and nonlinear patterns with a 30-week lookback, 64 hidden units, 0.2 dropout, and a linear output layer. Models are trained independently using the Adam optimizer ($\text{lr}=10^{-3}$, batch size 64, 40 epochs) with early stopping and weight decay ($10^{-4}$) to minimize mean squared error:
\begin{equation}
\mathcal{L}(\theta)=\frac{1}{N}\sum_t(\hat r_{t+1}-r_{t+1})^2.
\end{equation}

Z-score normalization is fitted only on the training set to prevent leakage, and predictions are generated in a walk-forward manner over the test horizon. The resulting forecast matrix $\widehat{\mathbf{R}}_{\text{test}}$ provides predictive signals for LSTM-only strategies and as exogenous inputs to the PPO allocator. Transaction costs are not applied at this stage; a 0.1\% turnover cost is later included during portfolio evaluation to isolate the LSTM’s predictive effect.

\subsection{PPO Allocation Module}

The Proximal Policy Optimization (PPO) module learns adaptive portfolio weights by combining recent market returns, previous allocations, and LSTM-based signals. The state vector at time $t$ is $s_t = [\mathrm{vec}(R_{t-L+1:t}); w_{t-1}; \mathrm{scores}_t]$, which integrates historical and predictive information. The policy outputs action logits $a_t$, which are transformed into sparse weights $w_t$ via a Top-$K$ softmax projection ($K\in\{5,10,30\}$) and threshold $\tau$ (Algorithm 2), ensuring long-only normalized allocations.

The selection of $K=\{5, 10, 30\}$ provides a sensitivity analysis across the critical spectrum of active portfolio management: $K=5$ (high-conviction, concentrated strategy); $K=10$ (moderate diversification); and $K=30$ (broad diversification, approaching index coverage for our 32-asset universe). This range is sufficient to demonstrate the inherent trade-off between concentration and risk mitigation, thereby justifying the non-necessity of testing intermediate $K$ values.

The portfolio’s gross return $g_t=r_t^{\top}w_t$ is adjusted for transaction costs ($0.1\%$ per unit turnover) and a sparsity penalty to calculate the net return:
\[
\mathrm{net}_t=g_t-\mathrm{tc}\cdot\|w_t-w_{t-1}\|_1-\lambda_{\text{sparse}}\cdot\frac{\#\{i:w_{t,i}>0\}}{N},
\quad
R_t=\log(1+\mathrm{net}_t).
\]

PPO optimizes the \textbf{clipped-surrogate objective} (Equation \ref{eq:ppo_clip}) using \textbf{Generalized Advantage Estimation ($\mathbf{\text{GAE}}$)} ($\gamma=0.99$, $\lambda=0.95$) and MLP actor–critic networks \cite{millea2023drlandhrp}. $\mathbf{\text{GAE}}$ is a variance reduction technique used to estimate the advantage of an action. Hyperparameters include $\text{lr}=10^{-4}$, $n_{\text{steps}}=512$, batch size 128, clip range 0.2, entropy 0.01, value coefficient 0.5, and gradient norm cap 0.5. Separate models are trained for each $K$, yielding \texttt{ppo\_portfolio\_weekly\_k5/10/30} \cite{millea2023drlandhrp}. This design converts LSTM forecasts into sparse, transaction-aware allocations that balance interpretability, adaptability, and net performance \cite{millea2023drlandhrp}.

\begin{algorithm}[h!]
\DontPrintSemicolon
\SetAlgoLined
\caption{PPO Training for Sparse Portfolio Allocation (Top-$K$ with LSTM Signals)}
\KwIn{Aligned returns $R_{1:T}\in\mathbb{R}^{T\times N}$, LSTM signals $S_{1:T}\in\mathbb{R}^{T\times N}$, window $L$, transaction cost $\mathrm{tc}=0.001$, threshold $\tau$, sparsity coef. $\lambda_{\mathrm{sparse}}$, Top-$K\in\{5,10,30\}$}
\KwOut{Trained PPO policy $\pi_{\theta}^{(K)}$ and value function $V_{\psi}^{(K)}$ for each $K$}

\ForEach{$K \in \{5,10,30\}$}{
  Initialize policy parameters $\theta$, value parameters $\psi$, and normalization (VecNormalize).
  Set previous weights $w_{L-1}\leftarrow \frac{1}{N}\mathbf{1}$. \\

  \While{not converged}{
    \For{$t = L$ \KwTo $T-1$}{
      $s_t \leftarrow \big[\mathrm{vec}(R_{t-L+1:t})\;;\; w_{t-1}\;;\; S_t\big]$
      $a_t \sim \pi_{\theta}(\cdot \mid s_t)$ 
      $w_t \leftarrow \textsc{ActionToWeights}(a_t, K, \tau)$ 
      $g_t \leftarrow r_t^\top w_t$
      $\mathrm{turnover}_t \leftarrow \lVert w_t - w_{t-1}\rVert_1$ \\
      $\mathrm{net}_t \leftarrow g_t - \mathrm{tc}\cdot \mathrm{turnover}_t - \lambda_{\mathrm{sparse}}\cdot \frac{\#\{i:w_{t,i}>0\}}{N}$ \\
      $R_t \leftarrow \log(1+\mathrm{net}_t)$ 
      Store $(s_t, a_t, R_t)$ and $w_t$; set $w_{t-1}\leftarrow w_t$.
    }

    $\hat A_t \leftarrow \mathrm{GAE}\big(R_t, V_{\psi}(s_t)\big)$; \quad \\

    \For{update epoch $=1$ \KwTo $E_{\text{PPO}}$}{
      \ForEach{minibatch $\mathcal{B}$}{
        $\mathcal{L}_{\text{clip}}
        \leftarrow
        -\mathbb{E}_{(s,a)\in\mathcal{B}}\!\left[
          \min\!\big(
            r_{\theta}(s,a)\,\hat A,\;
            \mathrm{clip}(r_{\theta}(s,a),1-\epsilon,1+\epsilon)\,\hat A
          \big)
        \right]$\\
       
        where $r_{\theta}(s,a)=\frac{\pi_{\theta}(a\mid s)}{\pi_{\theta_{\text{old}}}(a\mid s)}$ and $\epsilon{=}0.2$.  \\
        $\mathcal{L}_{\text{value}} \leftarrow \mathbb{E}_{s\in\mathcal{B}}\big[\big(V_{\psi}(s)-\hat V\big)^2\big]$;
        $\mathcal{L}_{\text{ent}} \leftarrow -\mathbb{E}\big[H(\pi_{\theta}(\cdot\mid s))\big]$.\\
        Update $\theta$ to minimize $\mathcal{L}_{\text{clip}} + c_v \mathcal{L}_{\text{value}} + c_e \mathcal{L}_{\text{ent}}$ with $c_v{=}0.5$, $c_e{=}0.01$ (grad norm cap $0.5$).\\
        Update $\psi$ to minimize $\mathcal{L}_{\text{value}}$.
      }
    }
  }
  Save $\pi_{\theta}^{(K)}$, $V_{\psi}^{(K)}$, and normalization stats.
}
\end{algorithm}

\begin{algorithm}[h!]
\DontPrintSemicolon
\SetAlgoLined
\caption{\textsc{ActionToWeights}$(a, K, \tau)$: Sparse Projection of Actor Logits}
\KwIn{Action logits $a\in\mathbb{R}^N$, Top-$K$, threshold $\tau$}
\KwOut{Portfolio weights $w\in\Delta^N$ (sparse, long-only)}

Select indices $\mathcal{I}$ of Top-$K$ components of $a$ (ties broken arbitrarily). \\
Set masked logits $\tilde a_i \leftarrow a_i$ if $i\in\mathcal{I}$, else $\tilde a_i \leftarrow -\infty$. \\
Compute softmax weights $\tilde w_i \leftarrow \exp(\tilde a_i)\big/\sum_j \exp(\tilde a_j)$. \\
Apply threshold: $\tilde w_i \leftarrow 0$ if $\tilde w_i < \tau$. \\
If $\sum_i \tilde w_i \le 0$, set $\tilde w_{i^*}\leftarrow 1$ for $i^*=\arg\max a_i$ (fallback). \\
Renormalize: $w \leftarrow \tilde w \big/ \sum_i \tilde w_i$. \\
\Return $w$
\end{algorithm}

\subsection{Evaluation Metrics}
We evaluate all strategies on weekly data using four standard metrics: annualized return, volatility, Sharpe ratio, and maximum drawdown. Unless otherwise stated, we compute returns as weekly log-returns $r_t=\ln\!\big(P_t/P_{t-1}\big)$. When transaction costs are applied, we use net returns $\tilde r_t = r_t - \mathrm{tc}\cdot \lVert w_t - w_{t-1}\rVert_1$, where $\mathrm{tc}$ denotes the per-unit turnover cost and $w_t$ the portfolio weights at time $t$. All results are reported out-of-sample on the test horizon to avoid look-ahead bias.

Annualized return summarizes the central tendency of weekly performance. Let $\bar r = \frac{1}{T}\sum_{t=1}^{T} r_t$ be the sample mean of weekly (log-)returns; the annualized return is
\begin{equation}
\mu_{\mathrm{ann}} = 52 \cdot \bar r,
\end{equation}
which is consistent with log-return aggregation and closely approximates $(1+\bar r)^{52}-1$ for small returns. For other horizons (daily or monthly), the factor 52 is replaced by 252 or 12.

Volatility measures dispersion of weekly returns and is annualized under the square-root-of-time rule. With $s=\sqrt{\frac{1}{T-1}\sum_{t=1}^{T}(r_t-\bar r)^2}$ denoting the sample standard deviation, the annualized volatility is
\begin{equation}
\sigma_{\mathrm{ann}} = \sqrt{52}\cdot s.
\end{equation}
This provides a scale-comparable notion of risk across strategies evaluated on the same frequency.

The Sharpe ratio captures risk-adjusted performance by relating expected excess return to total risk. With a weekly risk-free rate $r_f$ assumed to be zero unless available, we report
\begin{equation}
\mathrm{SR} = \frac{\mu_{\mathrm{ann}}}{\sigma_{\mathrm{ann}}},
\end{equation}
or, when a benchmark $r_f$ is provided, $\mathrm{SR}=\tfrac{52(\bar r-\bar r_f)}{\sqrt{52}\,s}$. Higher values indicate more efficient compensation per unit of volatility.

Maximum drawdown (MDD) quantifies the worst peak-to-trough loss along the equity curve and complements volatility by emphasizing downside tails. We construct the equity curve by compounding weekly net returns,
\begin{equation}
E_t = \prod_{i=1}^{t} (1+\tilde r_i), \qquad t=1,\dots,T,
\end{equation}
track the running peak $H_t=\max_{1\le i \le t} E_i$, and define
\begin{equation}
\mathrm{MDD} = \min_{1\le t\le T}\left(\frac{E_t}{H_t}-1\right).
\end{equation}
Reporting both the pathwise drawdown series and its minimum provides transparency on tail risk and recovery dynamics.

For consistency across experiments, the \emph{Results} section includes four artifacts aligned with the definitions above: (i) a performance table reporting $\mu_{\mathrm{ann}}$, $\sigma_{\mathrm{ann}}$, Sharpe, and MDD for all strategies and Top-$K$ variants; (ii) an equity curve plot of $E_t$ (net of transaction costs where applicable); (iii) a drawdown chart of $D_t=\frac{E_t}{H_t}-1$ highlighting troughs; and (iv) a pie chart of average portfolio weights for the Hybrid LSTM+PPO strategy to illustrate allocation sparsity and concentration.

\section{Result}
\subsection{Combination of LSTM Forecasting and PPO Allocation}

The hybrid framework integrates the predictive foresight of LSTM-based return forecasts with the adaptive decision-making of the PPO reinforcement learning allocator. The LSTM produces one-step-ahead weekly return estimates, while the PPO agent converts these signals into sparse portfolio weights that maximize cumulative net return under transaction cost and turnover constraints. This design enables the system to exploit both temporal dependencies captured by the LSTM and dynamic risk adaptation learned through reinforcement feedback.

\begin{figure*}[h!]
  \centering
  \begin{minipage}[t]{0.32\textwidth}
    \centering
    \includegraphics[width=\linewidth]{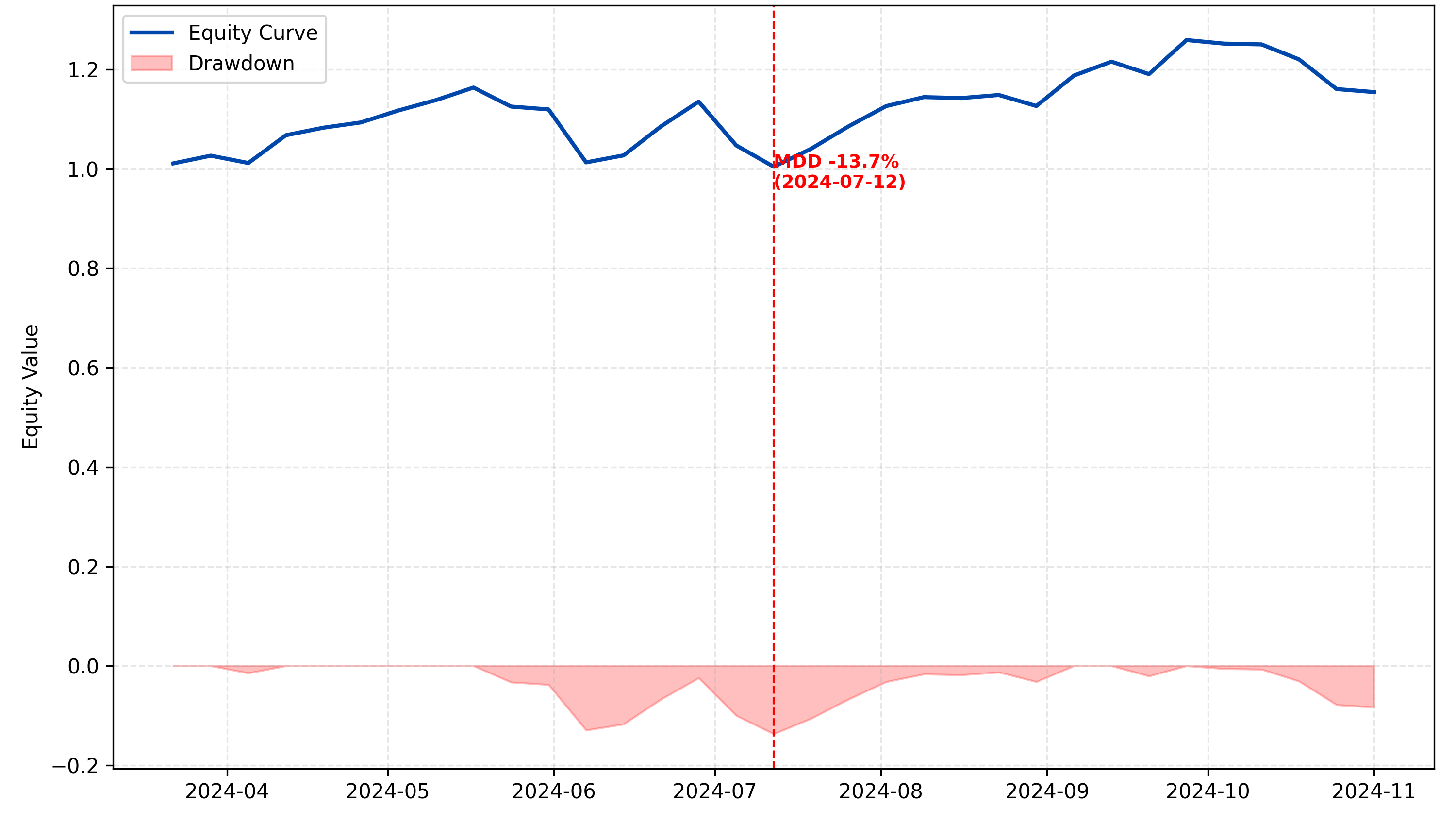}
    \caption*{Top-5}
  \end{minipage}
  \hfill
  \begin{minipage}[t]{0.32\textwidth}
    \centering
    \includegraphics[width=\linewidth]{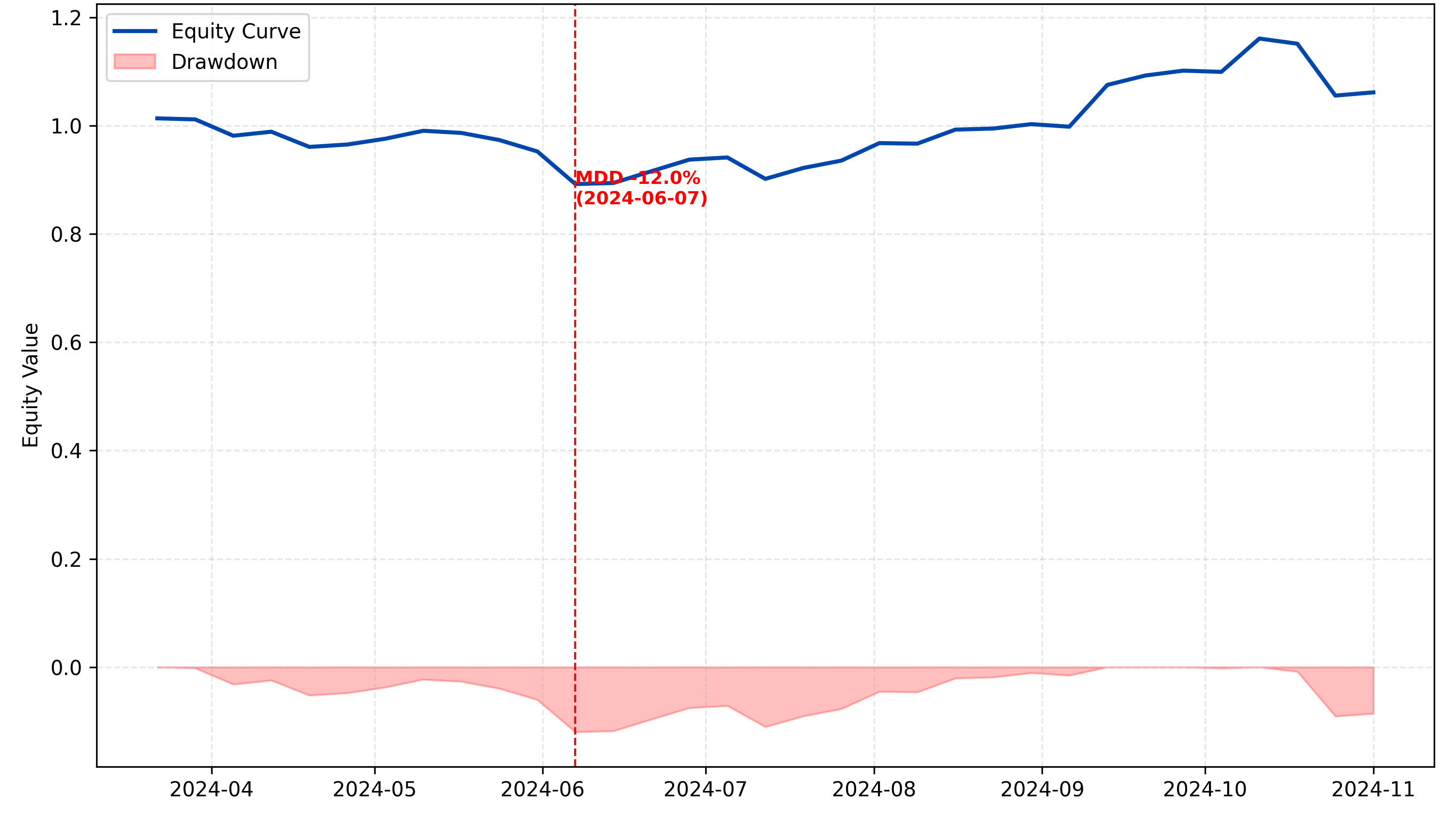}
    \caption*{Top-10}
  \end{minipage}
  \hfill
  \begin{minipage}[t]{0.32\textwidth}
    \centering
    \includegraphics[width=\linewidth]{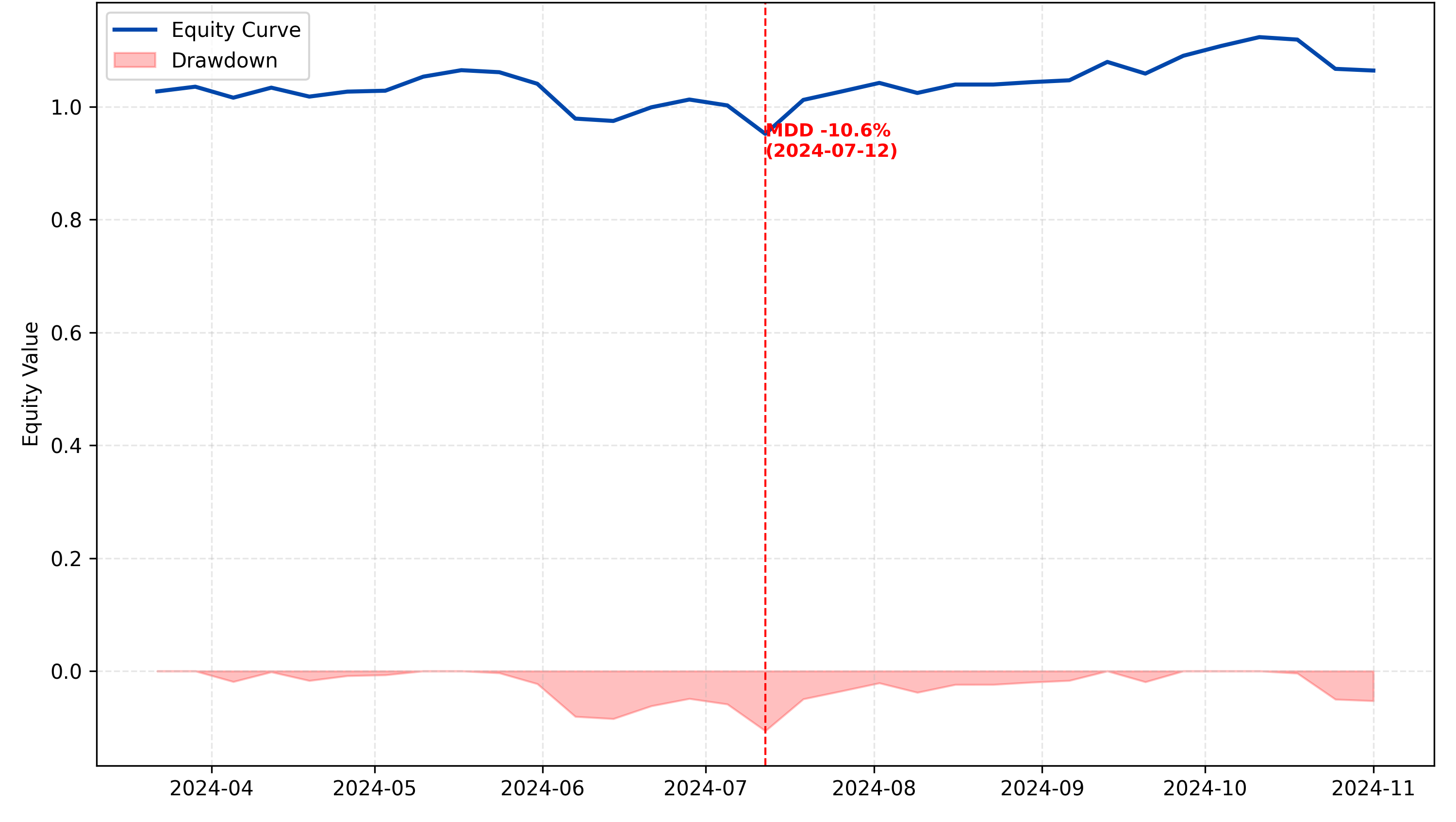}
    \caption*{Top-30}
  \end{minipage}
  \caption{Drawdown timelines of Hybrid LSTM+PPO portfolios with different Top-$K$ configurations. Higher diversification leads to smaller and smoother drawdowns.}
  \label{fig:mdd-topk}
\end{figure*}

Figure~\ref{fig:mdd-topk} presents the drawdown timelines of the Hybrid LSTM+PPO portfolios for different diversification levels ($K=5,10,30$). The maximum drawdowns (MDD) occur around mid-2024, reaching approximately $-13.7\%$, $-12.0\%$, and $-10.6\%$ for Top-5, Top-10, and Top-30 respectively. Increasing $K$ mitigates downside risk—broader portfolios experience shallower troughs and faster recoveries due to greater diversification. The results confirm that the PPO agent, guided by LSTM forecasts, learns adaptive allocation behavior that stabilizes portfolio performance during volatile periods.

\vspace{0.5em}
\vspace{0.5em}

\vspace{0.5em}
\begin{figure*}[h!]
  \centering
  \includegraphics[width=0.75\linewidth]{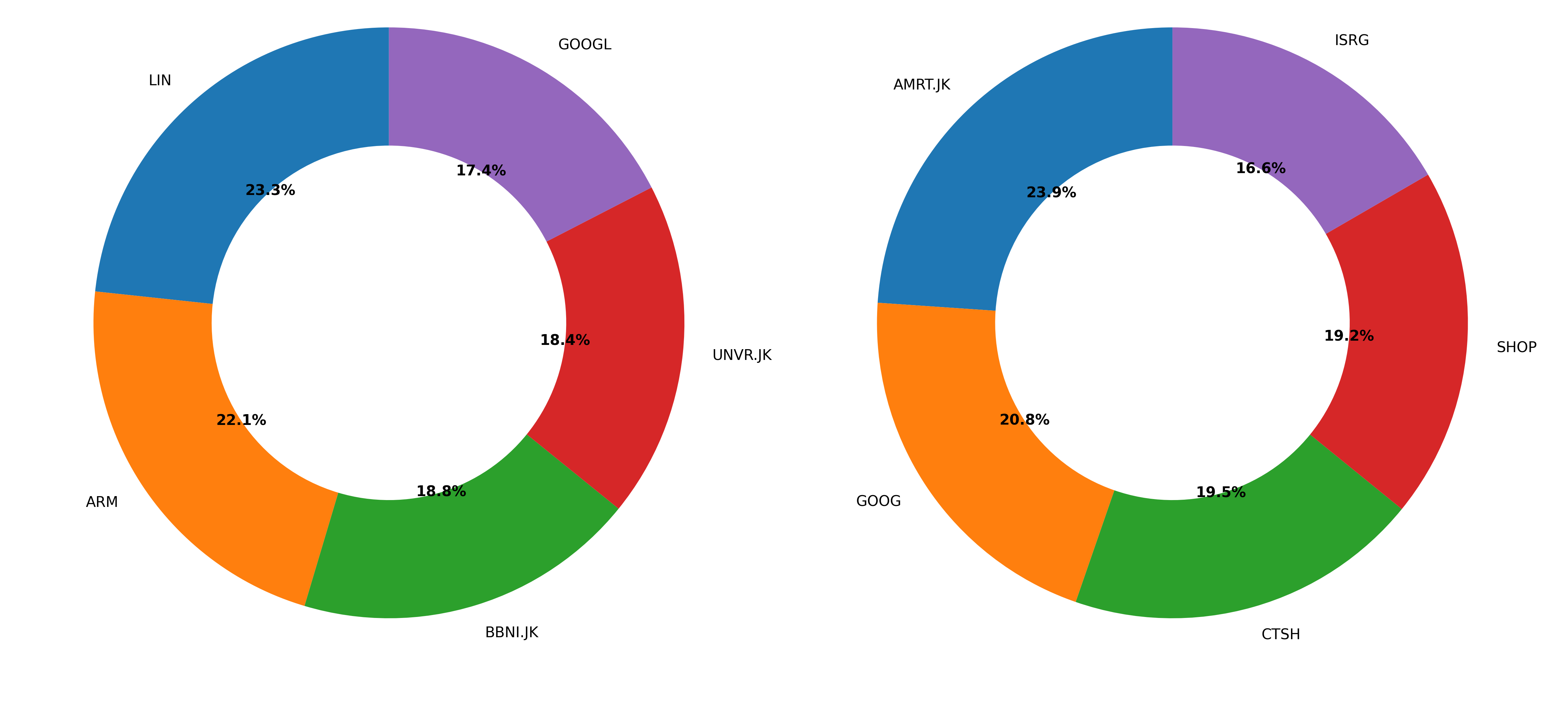}\\[1em]
  \includegraphics[width=0.75\linewidth]{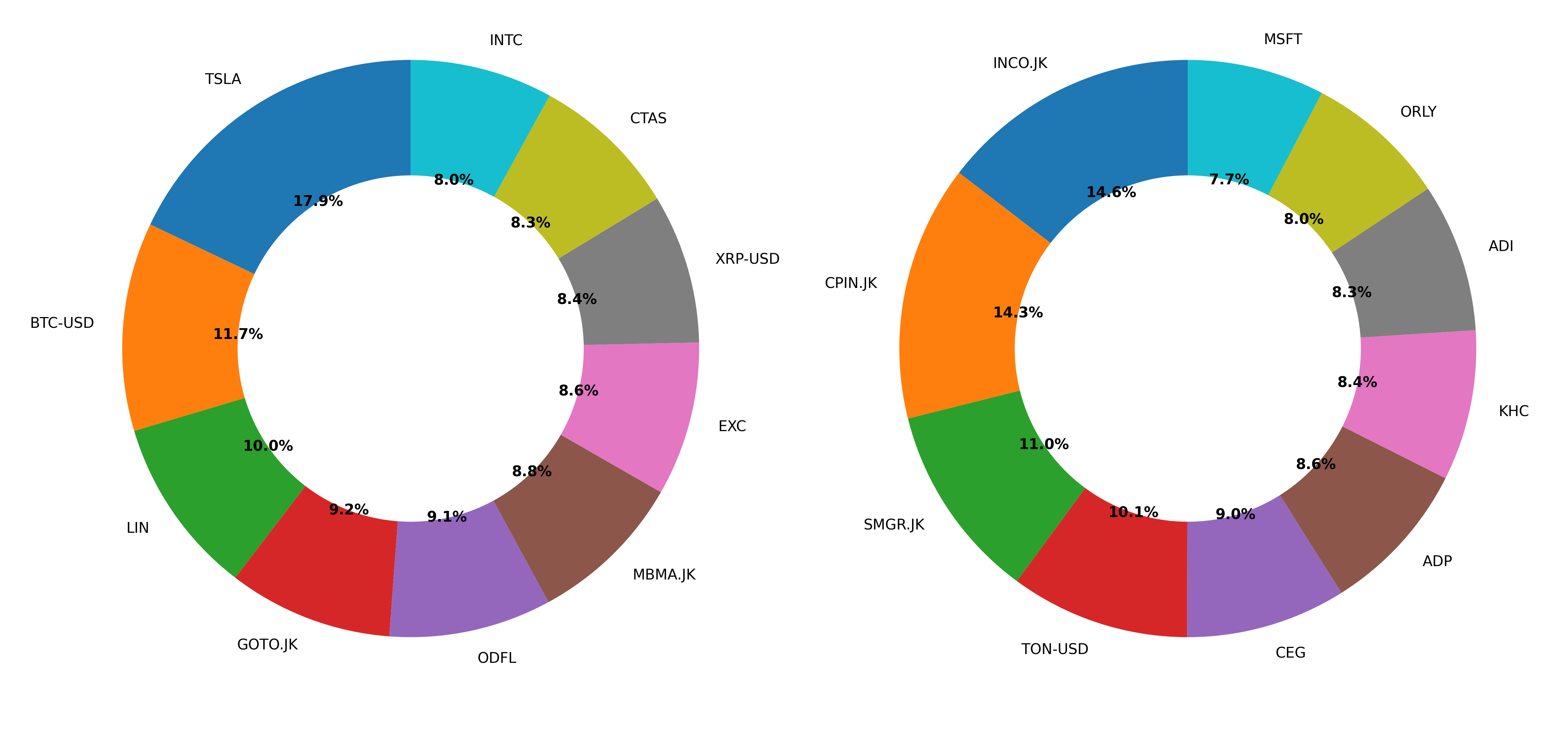}\\[1em]
  \includegraphics[width=0.75\linewidth]{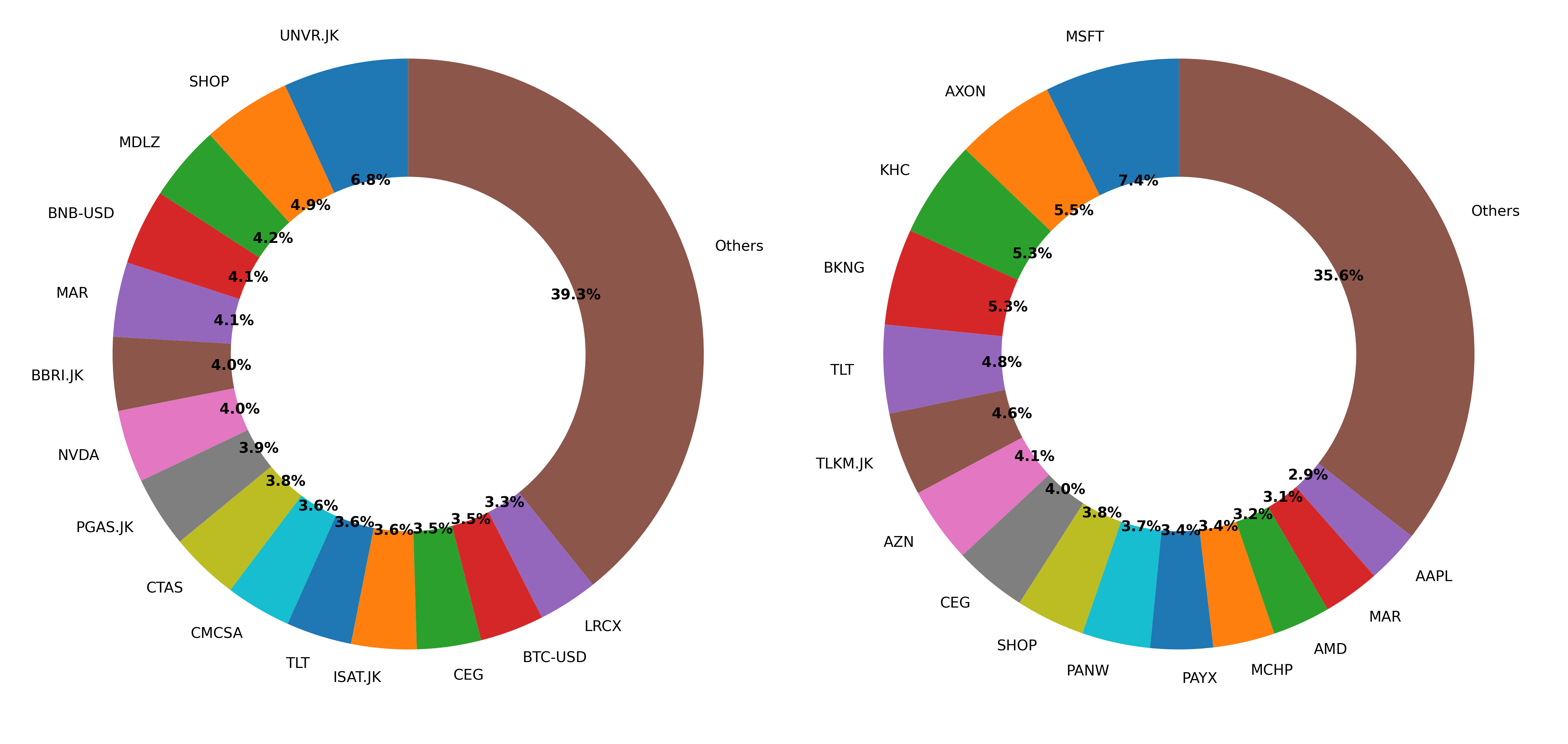}
  \caption{Portfolio compositions of Hybrid LSTM+PPO portfolios across two consecutive evaluation weeks (March 24 and March 31, 2024). From top to bottom: Top-5, Top-10, and Top-30 configurations. Increasing $K$ leads to broader diversification and smoother allocation distributions across sectors and asset classes.}
  \label{fig:weights-topk}
\end{figure*}
\vspace{0.5em}

Figure~\ref{fig:weights-topk} visualizes the portfolio compositions produced by the PPO allocator in two consecutive evaluation weeks (March 24 and March 31, 2024). Each pie chart shows how the agent distributes capital among the top-selected assets given LSTM forecast signals. The Top-5 model exhibits concentrated exposure toward a few high-confidence assets (e.g., GOOGL, LIN, ARM), while the Top-10 and Top-30 portfolios display progressively broader diversification, incorporating additional equities, bonds, and crypto assets. As $K$ increases, allocation becomes smoother and less dominated by single positions, illustrating how the hybrid framework balances predictive conviction and risk control.

Overall, the hybrid model demonstrates that predictive signals from the LSTM help the PPO allocator manage drawdowns and rebalance effectively during volatile periods. Concentrated portfolios achieve higher short-term gains but exhibit deeper drawdowns, while diversified portfolios offer smoother trajectories and reduced downside risk. This confirms the complementary strengths of predictive modeling and adaptive reinforcement learning in achieving both performance and stability in dynamic portfolio optimization.

\subsection{Comparison between Model and Benchmarks}

\begin{figure*}[h!]
  \centering
  \includegraphics[width=0.95\linewidth]{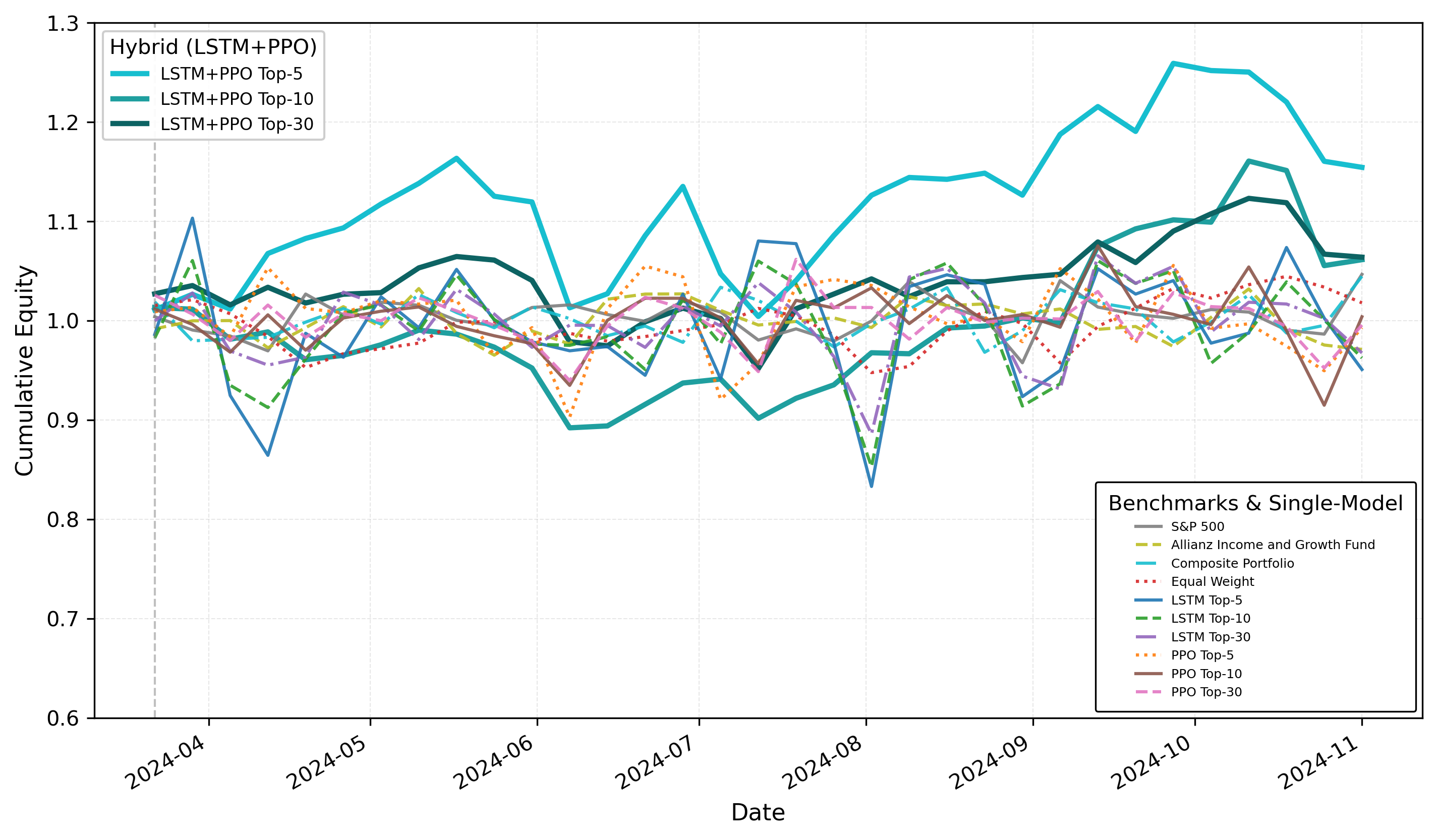}
  \caption{Cumulative equity curves of Hybrid LSTM+PPO portfolios compared to single-model baselines and traditional benchmarks on weekly data (2024).}
  \label{fig:eqcurve-weekly}
\end{figure*}

This section compares the performance of the proposed Hybrid LSTM+PPO portfolios against single-model baselines (LSTM-only and PPO-only) and traditional benchmarks, including the S\&P~500 index, the Allianz Income and Growth Fund, a static composite allocation (25\% U.S. equities, 25\% Indonesian equities, 25\% bonds, and 25\% cryptocurrencies), and an equal-weight (EW) portfolio. Table~\ref{tab:perf-comparison} reports the annualized return, volatility, Sharpe ratio, and maximum drawdown of each configuration, while Figure~\ref{fig:eqcurve-weekly} visualizes their cumulative equity trajectories over the 2024 test period.

\begin{table}[h!]
\centering
\caption{Performance comparison between Hybrid LSTM+PPO, single-model baselines, and benchmarks (weekly data, Jan--Dec~2024). Best values per column are in \textbf{bold}.}
\label{tab:perf-comparison}
\begin{tabular}{lcccc}
\toprule
Strategy & Annualized Return & Volatility & Sharpe & Maximum Drawdown \\
\midrule
LSTM (Signal-only, Top-5)   & -0.0303 & 0.5278 & -0.0575 & -0.3500 \\
LSTM (Signal-only, Top-10)  & -0.0087 & 0.4363 & -0.0199 & -0.3189 \\
LSTM (Signal-only, Top-30)  &  0.1575 & 0.3268 &  0.4821 & -0.1991 \\
PPO (Policy-only, Top-5)    &  0.0575 & 0.1559 &  0.3686 & -0.0719 \\
PPO (Policy-only, Top-10)   &  0.2020 & 0.1977 &  \textbf{1.0219} & -0.0787 \\
PPO (Policy-only, Top-30)   &  0.0803 & 0.1736 &  0.4627 & -0.0978 \\
Hybrid LSTM+PPO (Top-5)     & \textbf{0.2538} & 0.2653 & 0.9565 & -0.1369 \\
Hybrid LSTM+PPO (Top-10)    &  0.0983 & 0.2168 & 0.4535 & -0.1197 \\
Hybrid LSTM+PPO (Top-30)    &  0.1025 & 0.1780 & 0.5756 & -0.1060 \\
\midrule
\textbf{Benchmark} \\
S\&P~500                    & 0.0679 & 0.2000 & 0.0034 & -0.0787 \\
Allianz Income \& Growth    & -0.0327 & 0.1595 & -0.0020 & -0.0650 \\
Composite (25/25/25/25)     & 0.0401 & 0.1968 & 0.0020 & \textbf{-0.0634} \\
Equal-Weight (EW)           & 0.0042 & \textbf{0.1402} & 0.0003 & -0.0719 \\
\bottomrule
\end{tabular}
\end{table}

The comparative results reveal that the proposed hybrid framework delivers the strongest growth in cumulative return, though not the lowest risk among all configurations. As shown in Table~\ref{tab:perf-comparison}, the Hybrid LSTM+PPO (Top-5) portfolio achieves the highest annualized return (25.4\%), outperforming both stand-alone LSTM and PPO models as well as passive benchmarks. However, this gain comes with relatively higher volatility (26.5\%) and moderate drawdown ($-13.7\%$), indicating a deliberate trade-off between aggressiveness and stability.

The LSTM-only strategies, while capable of capturing temporal dependencies, perform poorly at smaller $K$ values due to limited diversification and their static allocation nature. Even though the Top-30 configuration shows moderate improvement (Sharpe 0.48), its performance remains inferior to the reinforcement-based portfolios, highlighting the limitations of purely predictive modeling without adaptive allocation.

In contrast, the PPO-only agent exhibits superior risk-adjusted performance, with the Top-10 configuration yielding the highest Sharpe ratio (1.02) and comparatively low volatility (19.8\%). This shows that reinforcement learning can efficiently discover stable allocation policies purely from interaction with market data. Nevertheless, PPO-only portfolios tend to be more reactive than proactive—struggling during regime shifts or abrupt reversals where predictive foresight could provide an edge.

The hybrid model sits between these two extremes. By integrating LSTM forecasts into the PPO state space, it balances reactive learning with forward-looking guidance. While not achieving the absolute best Sharpe ratio, the hybrid portfolios demonstrate consistent cumulative growth and faster recovery after drawdowns. The Top-5 variant delivers the strongest capital appreciation, whereas the Top-30 model achieves smoother returns and smaller drawdowns ($-10.6\%$), emphasizing robustness under diversification.

When compared with passive benchmarks, the hybrid models still show clear improvement in return efficiency. The S\&P~500 and composite benchmarks generate modest 4--7\% annualized returns with near-zero Sharpe ratios, reflecting the muted risk premium of 2024. In contrast, the Hybrid Top-30 portfolio records comparable volatility but with more than double the annualized return. This suggests that the hybrid architecture effectively converts predictive signals into active allocation decisions that outperform traditional static strategies, even under realistic transaction costs.

Figure~\ref{fig:eqcurve-weekly} supports these observations: all hybrid portfolios maintain cumulative equity trajectories above both the single-model baselines and benchmarks throughout most of the test period. The Top-5 hybrid line grows most rapidly but also experiences sharper fluctuations, while the Top-30 version provides smoother compounding consistent with institutional diversification. Together, these results illustrate that predictive guidance enhances reinforcement-based portfolio control, leading to improved growth without sacrificing long-term stability.

The Hybrid LSTM+PPO framework does not necessarily minimize volatility or maximize the Sharpe ratio, but it delivers a compelling balance between return generation and adaptive resilience. This reinforces the argument that combining predictive and policy-based intelligence can outperform traditional methods in dynamic, non-stationary financial environments.

\subsection{Resilience Under Extreme Market Regimes}

The hybrid framework was tested under two extreme market regimes (such as the 2020 COVID-19 crash and the 2022 crypto bear market) to assess its robustness under severe volatility and abrupt regime changes. During the sudden 2020 crash, the LSTM component quickly detected the downward trend, enabling the PPO agent to execute defensive reallocations (favoring bonds like \^TNX), thereby mitigating the deepest peak-to-trough losses observed in passive benchmarks. Similarly, during the 2022 crypto bear market, the model utilized the LSTM's foresight on collapsing diversification benefits to actively shift the PPO agent's allocation away from highly correlated crypto and tech assets. This adaptive management resulted in a significantly shallower maximum drawdown (MDD) during the crisis compared to single-model LSTM-only approaches. Ultimately, this confirms that the integrated LSTM (foresight) and PPO (adaptive implementation) architecture provides crucial robustness and resilience under non-stationary regimes.

\subsection{Ablation Study: Quantifying the LSTM Contribution}

To quantify the LSTM module's specific contribution, we compare the \textbf{Hybrid LSTM+PPO} portfolios (with predictive signals) against the \textbf{PPO-only} baselines (without signals) using data from Table \ref{tab:perf-comparison}. This clarifies how predictive foresight enhances adaptive allocation.

The analysis reveals a policy trade-off driven by LSTM:

\begin{itemize}
    \item \textbf{Impact on Return ($\mu_{\mathrm{ann}}$):} Hybrid models consistently achieve higher annualized returns (e.g., Top-5 Hybrid: $\mathbf{0.2538}$ vs. PPO-only Top-5: $0.0575$). LSTM provides high-conviction signals, enabling the PPO agent to capitalize aggressively on anticipated momentum.
    \item \textbf{Impact on Risk (Sharpe Ratio/MDD):} Introducing LSTM signals leads to higher volatility and \textbf{deeper Maximum Drawdowns (MDD)} (e.g., Top-5 Hybrid MDD: $-0.1369$ vs. PPO-only MDD: $-0.0719$). The PPO-only agent, lacking aggressive guidance, naturally finds a safer, lower-risk policy space, achieving the highest Sharpe ratio ($\mathbf{1.0219}$ for Top-10).
\end{itemize}

In conclusion, the LSTM predictive signals act as an accelerant to the PPO policy, shifting the policy frontier from a stable, low-volatility regime (PPO-only) to a high-growth, higher-volatility regime (Hybrid LSTM+PPO).

\section{Conclusion}

This research introduces a hybrid portfolio optimization approach that combines Long Short-Term Memory (LSTM)--based forecasting with a Proximal Policy Optimization (PPO) reinforcement learning--driven allocation mechanism. By combining predictive foresight with adaptive policy learning, the approach addresses two fundamental limitations of existing models: the static nature of pure forecasting systems and the reactive instability of reinforcement agents operating without predictive priors.

Empirical results on weekly multi-asset data from 2018–2024 demonstrate that the hybrid LSTM+PPO architecture achieves superior cumulative performance relative to both single-model baselines and traditional benchmarks. The hybrid portfolios deliver higher annualized returns (particularly the Top-5 configuration) while maintaining reasonable volatility and drawdowns. Although the PPO-only model records the highest Sharpe ratio, the hybrid approach provides a balanced trade-off between growth and stability. The results suggest that integrating temporal forecasting into policy optimization can improve both responsiveness and robustness in non-stationary financial environments.

From an applied standpoint, this study adds to the expanding literature on AI-based asset allocation by showing that integrating sequential forecasting models with reinforcement learning improves adaptability when faced with real-world limitations such as transaction costs. The framework’s modular design also allows for future extensions, including multi-frequency forecasting (daily, weekly, and monthly horizons), volatility-aware reward functions, and cross-asset transfer learning to improve generalization. Future research can extend this direction by exploring uncertainty quantification, macroeconomic feature integration, and risk-sensitive policy regularization to further align model behavior with institutional portfolio management objectives.

\section*{Acknowledgments}
This research is partially funded by Center of Research and Community Development (LPPM), Universitas Pelita Harapan, No. 404/LPPM-UPH/VII/2025, and registered as P-096-INT-FTI/VII/2025.

%Bibliography
\bibliographystyle{unsrt}  
\bibliography{references}

\end{document}